# Robust Image Processing Techniques for Construction Environment Monitoring Using Underwater Robots

Seunghee Yun[1*], Geonmo Yang[1*], Juhui Lee[1], Changbeom Park[2], Jeahyung Choi[2], and Younggun Cho[3†]

[1]Department of Electrical and Computer Engineering, Inha University
[2]Hyundai Engineering & Construction Co., Ltd
[3]Department of Electrical and Electronic Engineering, Inha University

**Abstract:** This paper proposes a robust image processing framework for underwater robot-based construction environment monitoring, targeting complex degradations observed in real marine environments. Unlike conventional approaches that mainly consider absorption and backscattering, real underwater imagery is strongly affected by depth-dependent forward scattering blur and particle-induced degradations such as marine snow. To address this, we introduce a staged processing pipeline that sequentially models background degradation via depth-aware forward scattering and foreground degradation using realistic marine snow patterns extracted from real images. The resulting synthetic data are used to retrain an existing Joint-ID network without modifying its architecture, enabling an isolated evaluation of dataset realism. In addition, a lightweight post-processing scheme is applied to enhance contrast and structural clarity. Experiments on real underwater datasets collected in Korean coastal environments demonstrate consistent improvements in visual quality and UIQM scores. The results indicate that explicitly modeling forward scattering and realistic particle effects effectively reduces the synthetic-to-real gap and improves practical applicability in real-world underwater robotic operations.

## I. Introduction

In recent years, underwater robots have been increasingly employed in various fields such as marine surveys, inspection of subsea structures, and underwater construction and maintenance [1,2]. However, underwater images acquired in real sea environments suffer from severe quality degradation due to the optical characteristics of the underwater medium, which impairs structure identification and monitoring performance [3,4,5]. In particular, real coastal images exhibit not only color attenuation and contrast reduction but also edge blurring caused by forward scattering and particle-induced occlusion caused by marine snow, which makes visibility and the interpretation of structural information even more difficult.

Existing underwater image processing and enhancement techniques have proposed various approaches to mitigate the image degradation that occurs in underwater environments [6,7]. However, most studies have mainly focused on restoring the degraded sharpness of underwater images and correcting color distortion [8,9], while the design of synthetic datasets reflecting the complex degradations of real underwater environments has been addressed only to a limited extent. Consequently, many studies rely on experiments based on water-tank environments or simplified synthetic data [10], and thus fail to sufficiently reflect the complex image degradation characteristics observed in underwater robot operations in real sea environments [1,11]. In particular, as shown in Fig. 1, existing synthetic underwater images do not adequately reproduce the particle distribution and density characteristics of marine snow observed in real sea environments. Such discrepancies are not merely visual mismatches; they lead to distribution gaps between training data and real-world data, limiting the ability of restoration models to reliably handle the particle-induced degradation and structural edge deterioration that appear in real sea images [12]. This instability of structural information is a major factor that prevents the image quality enhancement performance required in real underwater robot operating environments from being sufficiently achieved.

This study defines the image processing requirements applicable to underwater robot operations under the premise of the degradations occurring in real underwater environments, and presents an approach that reflects these requirements. To this end, we utilized image data directly acquired by an underwater robot in Korean coastal waters with poor visibility, and comprehensively considered forward scattering, realistic marine snow distributions, and the preservation of contrast and structural information. The contributions of this study are as follows.

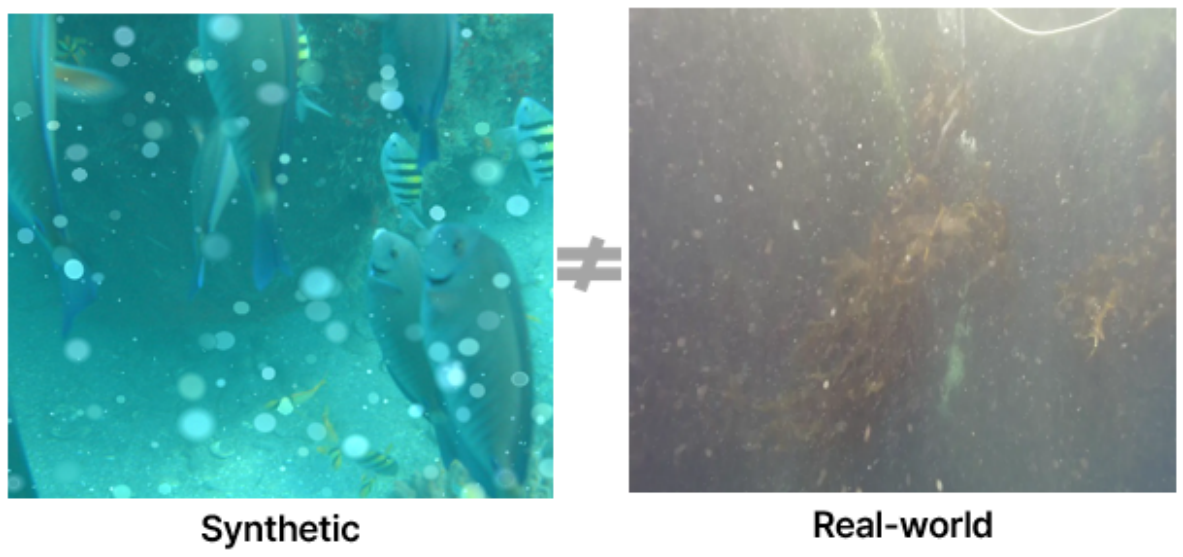

Fig. 1. Structural limitations of synthetic data vs. real-world data

First, we propose an image processing technique that accounts for the forward scattering, marine snow, and contrast degradation occurring in real underwater robot operating environments.

---

[*] Represents equal contribution
[†] Corresponding Author

Seunghee Yun[1*]: Graduate Student, Department of Electrical and Computer Engineering, Inha University (shyun28@inha.edu, ORCID 0009-0002-0736-0976)
Geonmo Yang[1*]: Graduate Student, Department of Electrical and Computer Engineering, Inha University (ygm7422@inha.edu, ORCID 0009-0009-1273-9631)
Juhui Lee[1]: Graduate Student, Department of Electrical and Computer Engineering, Inha University (dlwngml6635@inha.edu, ORCID 0009-0000-9096-8805)
Changbeom Park[2]: R&D Division, Hyundai Engineering & Construction (cbpark@hdec.co.kr, ORCID 0009-0008-7641-8346)
Jeahyung Choi[2]: R&D Division, Hyundai Engineering & Construction (jhchoi@hdec.co.kr, ORCID 0009-0005-6430-3574)
Younggun Cho[3†]: Professor, Department of Electrical and Electronic Engineering, Inha University (yg.cho@inha.ac.kr, ORCID 0000-0003-2025-7770)

※ This research was supported by the Korea Heritage Service and the National Research Institute of Cultural Heritage in 2026 (Project No.: RS-2026-25518173), by the Institute of Information & Communications Technology Planning & Evaluation (IITP) grant funded by the Korea government (MSIT) (No. 2022-0-00448/RS-2022-II220448), and by Hyundai Engineering & Construction

Second, we apply the proposed technique to image data acquired by an underwater robot in real Korean coastal waters and validate its performance through qualitative and quantitative evaluations.

## II. Related Works

2.1 Physics-based Synthesis

Image degradation in underwater environments is physically described based on the Jaffe-McGlamery image formation model, and numerous studies have generated synthetic training data based on it. Akkaynak and Treibitz [4] experimentally demonstrated that the attenuation coefficient of the direct transmission path and that of the backscattering path are independent physical quantities, and proposed a revised underwater image formation model distinct from conventional atmospheric scattering models. Li et al. [13] generated a large-scale synthetic paired underwater dataset by applying ten wavelength-dependent attenuation coefficients according to the Jerlov water type classification to NYU-v2 depth images, and more recently, Kaneko et al. [14] presented the PHISWID dataset, which combines a color degradation model that jointly considers the downwelling depth and the line-of-sight distance with Jaffe-McGlamery light-scattering-based marine snow synthesis. Joint-ID [6], on which this study is based, also adopts a physics-based synthesis pipeline, but it has limitations in explicitly modeling depth-dependent forward scattering blur and marine snow at the same time.

2.2 GAN-based Synthesis

Approaches that use generative adversarial networks (GANs) to translate terrestrial images into underwater images, or to synthesize underwater degradation patterns in an unsupervised manner, have also been actively studied. WaterGAN by Li et al. [10] synthesized underwater images from terrestrial RGB-D images through a physics-inspired generator with a sequential attenuation-scattering-camera model structure, and Fabbri et al. [15] generated underwater-terrestrial image pairs using CycleGAN-based unsupervised domain translation. In addition, Galetto and Deng [16] proposed a method for independently synthesizing only marine snow particles with a GAN. However, GAN-based approaches are vulnerable to mode collapse, and since their generation process relies on implicit data-dependent mappings, they have the inherent limitation that individual degradation factors such as forward scattering and marine snow are difficult to interpret physically or to control independently.

2.3 Diffusion-based Degradation Synthesis

With the recent advancement of diffusion models, attempts to leverage them for image synthesis have been reported in various fields. In the underwater imaging domain, Zhang et al. [17] proposed Atlantis, which combines ControlNet and Stable Diffusion to generate underwater-style images from terrestrial depth maps, and in the general imaging domain, Rajagopalan et al. [18] presented GenDeg, which synthesizes various degradations such as haze, rain, and low light with text conditioning based on diffusion. However, these studies focus on scene-level style transfer or general atmospheric degradations, and do not individually model underwater-specific physical degradation factors such as wavelength-dependent attenuation, forward scattering, and marine snow. Moreover, diffusion-based approaches have fundamental limitations for underwater degradation synthesis, including the high computational cost of iterative sampling, the inability to independently control specific degradation factors, and the difficulty of physical interpretation due to their black-box nature; their application to synthesis studies that separate and control physical degradation factors thus appears to remain very limited.

Taking these observations together, considering the applicability in real underwater working environments and the controllability of the data generation process, this study adopts a relatively lightweight synthesis scheme in which forward scattering and marine snow can be explicitly controlled.

## III. Preliminary

This section provides the common preliminaries for describing the image processing techniques proposed in the subsequent sections. This study utilizes a physics-based synthetic underwater image model [6], and this section focuses on the composition and characteristics of the corresponding synthesis process.

The synthetic underwater images used in this study are generated from clean RGB–D images acquired in terrestrial environments. Given an input clean image $J_c(x)$ and the corresponding per-pixel distance information $d(x)$, for channel $c \in \{R, G, B\}$, the basic synthetic underwater image $I_c^{basic}(x)$ is defined as follows.

$$I_c^{basic}(x) = J_c(x) \cdot T_c\big(d(x)\big) + B_c^{\infty}\Big(1 - T_c\big(d(x)\big)\Big) \quad (1)$$

The above equation simultaneously models the attenuation of light passing through the underwater medium and the mixing of ambient light. Here, $T_c(d) = \exp(-\beta_c \cdot d)$ is the per-channel transmission based on the Beer–Lambert law, which represents the degree to which the original color component is attenuated with depth $d$. $\beta_c$ is the attenuation coefficient that varies with water type and wavelength, and $B_c^{\infty}$ denotes the homogeneous ambient light component present in the underwater environment. The synthesis process reflects the absorption and backscattering effects of the underwater medium, and as a result, the color distortion and global contrast reduction observed in real underwater images are naturally reproduced.

In addition, to reflect the color attenuation characteristics under various water conditions, this study samples $\beta_c$ and $B_c^{\infty}$ based on the Jerlov water type classification. This makes it possible to simulate different underwater optical characteristics and, as a result, to construct basic synthetic underwater images that comprehensively reflect the characteristics of underwater images observed in real sea environments.

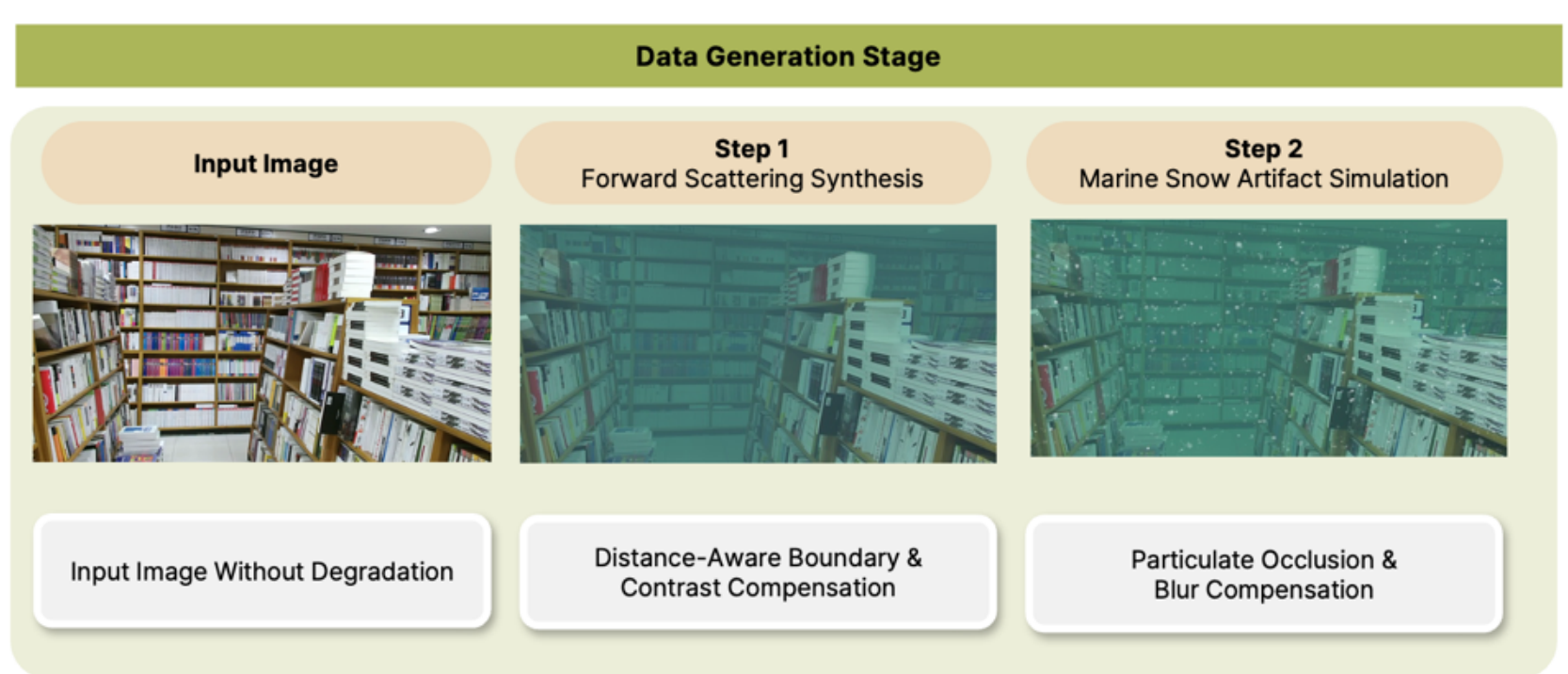

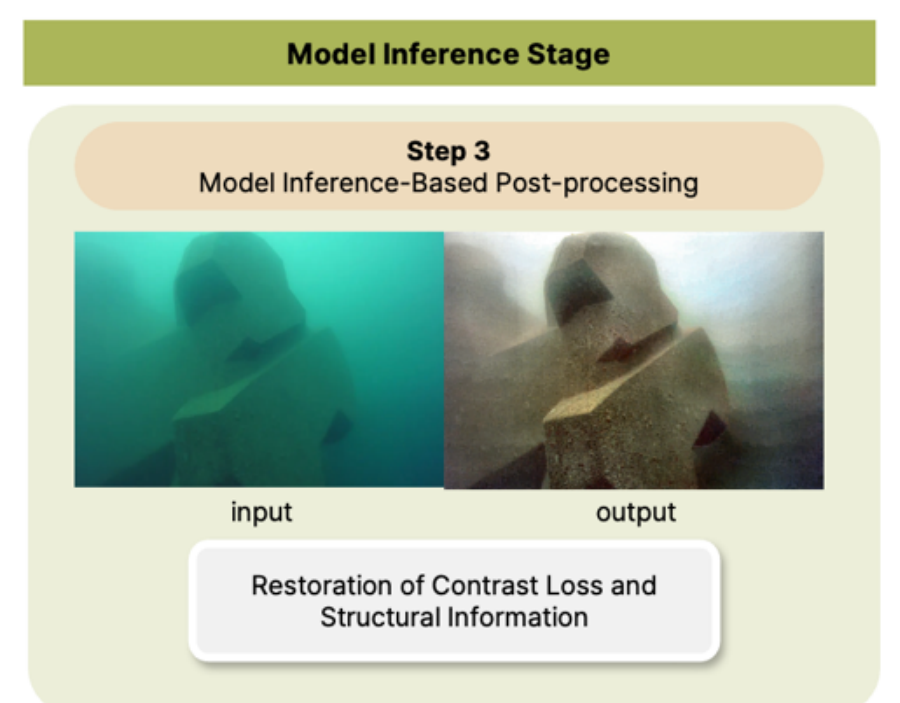

Fig. 2. Synthetic data generation stages and model inference–based post-processing pipeline

## IV. Method

This section describes the image processing techniques designed to progressively account for the major degradation factors that appear in images from underwater robots operating in real underwater environments. The proposed method is structured to sequentially consider background degradation caused by the underwater medium, background degradation caused by suspended particles, foreground degradation caused by suspended particles, and the resulting loss of contrast and structural information. An overview of the proposed overall system is presented in Fig. 2.

### 4.1 First-stage Synthesis Based on Forward Scattering

Conventional synthetic underwater images are generated considering only the color attenuation caused by absorption and backscattering, and therefore have the limitation that the boundaries of distant structures remain excessively sharp compared with real underwater environments. In contrast, in real underwater environments, the effect of forward scattering by fine particles in the underwater medium causes visibility to deteriorate and structural boundaries to gradually diffuse as depth increases. This distance-dependent boundary weakening is one of the major factors that impair structure identifiability in real underwater working environments.

To reflect these characteristics, this study performs a first-stage reproduction of the visibility degradation and boundary diffusion caused by forward scattering by applying blur that increases with depth. According to Jaffe-McGlamery-type underwater image formation models, the forward scattering component can be approximated by applying a distance-dependent point spread function (PSF) to the direct component [19]. Moreover, since small-angle scattering is dominant in underwater environments, the spatial diffusion caused by forward scattering can be effectively approximated by a Gaussian-shaped PSF [20]. Accordingly, for channel $c \in \{R, G, B\}$, this study defines a two-dimensional Gaussian PSF whose standard deviation varies with depth $d(x)$ as follows.

$$K_{fs,c}\big(u; d(x)\big) = \frac{1}{2\pi\sigma_c^2\big(d(x)\big)} \exp\left( -\frac{\| u \|^2}{2\sigma_c^2\big(d(x)\big)} \right) \quad (2)$$

Here, $u$ denotes the spatial coordinate offset from the center of the PSF.

The standard deviation $\sigma_c(d)$ is set as follows to reflect the blur characteristic that increases with depth.

$$\sigma_c(d) = \sigma_{0,c} + k_c \cdot d \quad (3)$$

Here, $\sigma_{0,c}$ is the minimum blur size in the near range, and $k_c$ is the blur growth rate with increasing depth. To concisely reflect the tendency of the forward scattering effect to increase gradually with distance, this study approximates $\sigma_c(d)$ as a linear function of depth. The specific values of $\sigma_{0,c}$ and $k_c$ are presented in Section 5.1.

Since the defined PSF is a spatially varying filter whose kernel shape changes with depth $d(x)$, the ideal forward scattering blurred image $\tilde{I}_c^{fs}(x)$ is defined as follows.

$$\tilde{I}_c^{fs}(x) = \int K_{fs,c}\big(x - x'; d(x)\big)\, I_c^{basic}(x')\, dx' \quad (4)$$

Here, $I_c^{basic}(x)$ is the basic synthetic underwater image obtained from Eq. (1). This equation represents a spatially varying blur in which different blur kernels are applied depending on depth.

However, since directly applying the spatially varying PSF to every pixel is computationally expensive, this study first generates a fixed-blur image $\bar{I}_c^{fs}(x)$ that reflects the maximum blur level, and then approximates spatially varying forward scattering by blending it with the basic synthetic image using a depth-dependent weighting function. The final forward-scattering image $I_c^{fs}(x)$ is defined as follows.

$$I_c^{fs}(x) = \Big(1 - \alpha_{fs}\big(d(x)\big)\Big)\, I_c^{basic}(x) + \alpha_{fs}\big(d(x)\big)\, \bar{I}_c^{fs}(x) \quad (5)$$

Here, $\alpha_{fs}(d) \in [0,1]$ is a weighting function that increases monotonically with depth, and it is defined in this study as follows.

$$\alpha_{fs}(d) = 1 - exp(-\gamma d) \quad (6)$$

Here, $\gamma > 0$ is a parameter that controls the rate at which the blur is applied. At close range, $\alpha_{fs}(d)$ is small, so the structure of the basic synthetic image is preserved, whereas with increasing distance, $\alpha_{fs}(d)$ increases, so that the boundary diffusion caused by forward scattering is progressively reflected.

Meanwhile, in the approximation process of Eq. (5), the forward scattering blur is applied to the entirety of $I_c^{basic}(x)$. Although forward scattering mainly corresponds to the direct component in a strict physical model, this study adopts this approximate implementation to efficiently reproduce the distance-dependent boundary weakening and visibility degradation observed in real underwater images. In addition, since the basic synthetic image in Eq. (1) already includes absorption and backscattering, separating and processing the direct and backscattering components again would considerably increase the implementation complexity. Therefore, this study applies the spatially varying blur approximation to the entirety of $I_c^{basic}(x)$.

Stage 1 of the data generation process in Fig. 2 illustrates the result of applying the first-stage forward-scattering-based synthesis to the same input image. After forward scattering is applied, structural boundaries are progressively softened in more distant regions, and the distance-dependent boundary diffusion and visibility degradation observed in real underwater images are reproduced more closely.

4.2 Incorporation of Marine Snow Artifacts and Retraining-based Dataset Application

Images acquired in real underwater environments frequently exhibit the marine snow phenomenon caused by suspended particles, which is a major degradation factor that induces particle-induced occlusion and local light diffusion in the foreground region [11, 16].

However, in existing synthetic underwater data, marine snow is often treated as simple random noise or generated in forms that differ from the particle size, density, and spatial distribution characteristics observed in real sea environments. Such simplified modeling causes a distribution mismatch between synthetic and real sea data, which can lead to performance degradation when the trained network is applied to real underwater robot environments.

To alleviate this problem, this study adds particle-induced foreground degradation to the synthetic images based on marine snow patterns extracted from real underwater images [16]. Specifically, the marine snow patterns are constructed as follows. First, a real underwater image is converted to grayscale, and locally bright particle regions are separated based on adaptive thresholding. Morphological refinement (opening and closing) is then applied to refine the particle regions from the background structures and to generate a binary mask $B(x)$. The color components of the marine snow $M_c(x)$ directly take the per-channel pixel values of the original image at locations where $B(x) = 1$, thereby reflecting the brightness and color distribution of real particles. Fig. 3 shows an example of a marine snow mask extracted through this process.

Marine snow is modeled as a foreground layer separated from the forward-scattering background image $I_c^{fs}(x)$ and is composited via alpha blending. For channel $c \in \{R, G, B\}$, the final synthetic image incorporating marine snow $I_c^{ms}(x)$ is defined as follows.

$$I_c^{ms}(x) = (1 - \alpha_{ms}(x)) I_c^{fs}(x) + \alpha_{ms}(x) M_c(x) \qquad (7)$$

Here, $\alpha_{ms}(x) \in [0,1]$ is the marine snow alpha map defined according to particle positions and sizes, and it is generated by applying, to the binary mask $B(x)$, Gaussian smoothing with standard deviation $\rho$.

$$\alpha_{ms}(x) = \left(G_\rho * B\right)(x) \qquad (8)$$

Since the input to the Gaussian smoothing is a binary mask taking values in $\{0, 1\}$ and the kernel is normalized, the output naturally lies within the range $[0, 1]$. The resulting $\alpha_{ms}(x)$ reflects a gradual light diffusion effect at the particle edges, and realistically simulates the irregular spatial distribution of particles and the variation in occlusion strength.

To validate the effectiveness of the proposed synthetic dataset, this study adopts the Joint-ID [6] model as the baseline model for analysis. The overall network architecture of Joint-ID is presented in Fig. 4. To clearly analyze the influence of dataset design, we keep the network architecture and loss function of Joint-ID unchanged and perform retraining by changing only the training data. Specifically, by replacing the existing synthetic dataset with the proposed dataset that incorporates forward scattering and real-image-based marine snow artifacts, we quantitatively and qualitatively evaluate how differences in synthetic data composition affect underwater image restoration performance.

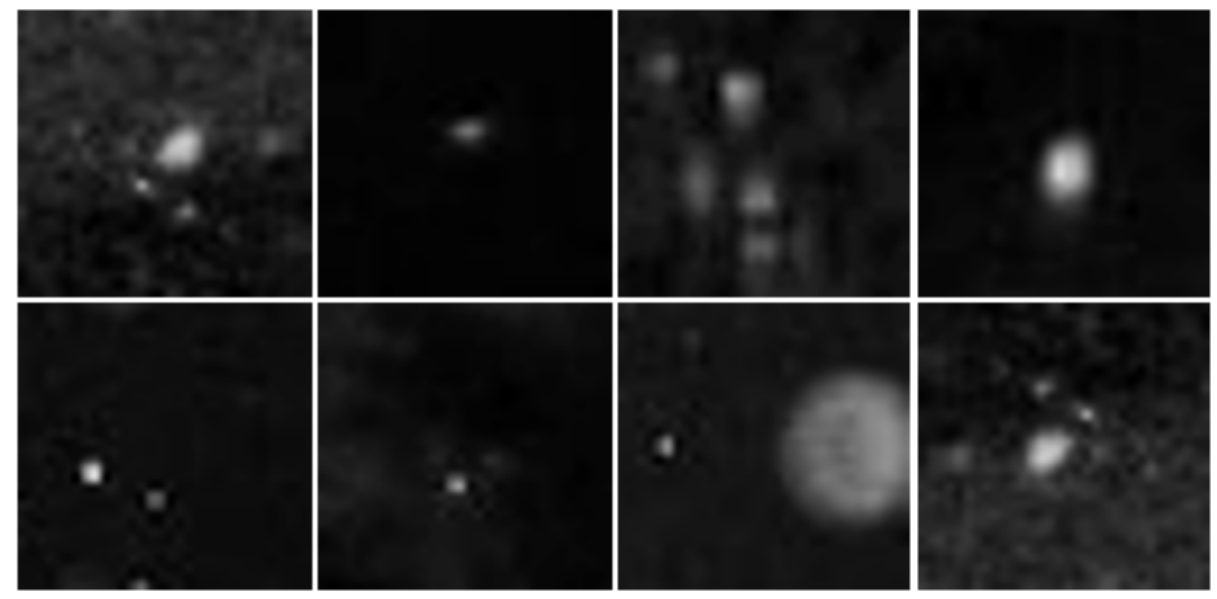

Fig. 3. Example of a marine snow mask image

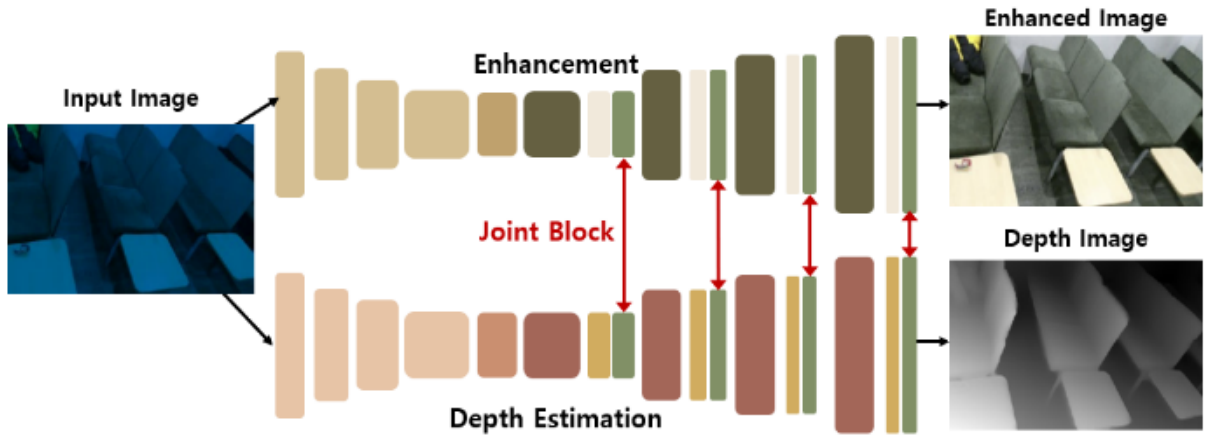

Fig. 4. Architecture of the Joint-ID network

4.3 Post-processing for Preserving Contrast and Structural Information

Although images containing forward scattering and marine snow reflect real underwater environments, they may cause contrast reduction and weakening of structural boundaries. To compensate for these problems, this study applies a histogram-equalization-based post-processing technique after the network output [21]. This technique is applied to the output image without additional training or changes to the network architecture, and it readjusts the overall intensity distribution of the image through histogram equalization.

Then, to conservatively emphasize the boundary information of structures,

a Laplacian of Gaussian (LoG) filter is combined with the histogram-equalized image $I^{HE}$. The final post-processed output image $I^{out}$ is defined as follows.

$$I^{out}(x) = I^{HE}(x) - \lambda(\nabla^2 G_\sigma * I^{HE})(x) \qquad (9)$$

Here, $\nabla^2 G_\sigma$ is the Laplacian of a Gaussian function with standard deviation $\sigma$, and $\lambda > 0$ is a weight that controls the strength of edge enhancement. This operation serves to complementarily emphasize local intensity variations, such as structural boundaries, in underwater images in which low-frequency components are dominant.

Such a post-processing stage compensates for the contrast degraded by forward scattering and particle-induced degradation, and reveals the boundary information of structures more reliably, thereby improving visual interpretability.

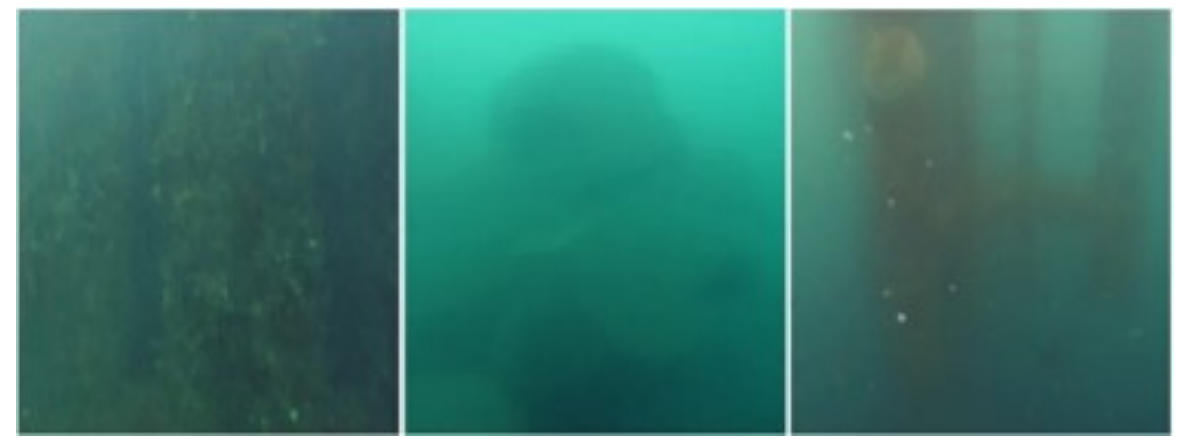

Fig. 5. Underwater structures acquired in real coastal waters of Korea

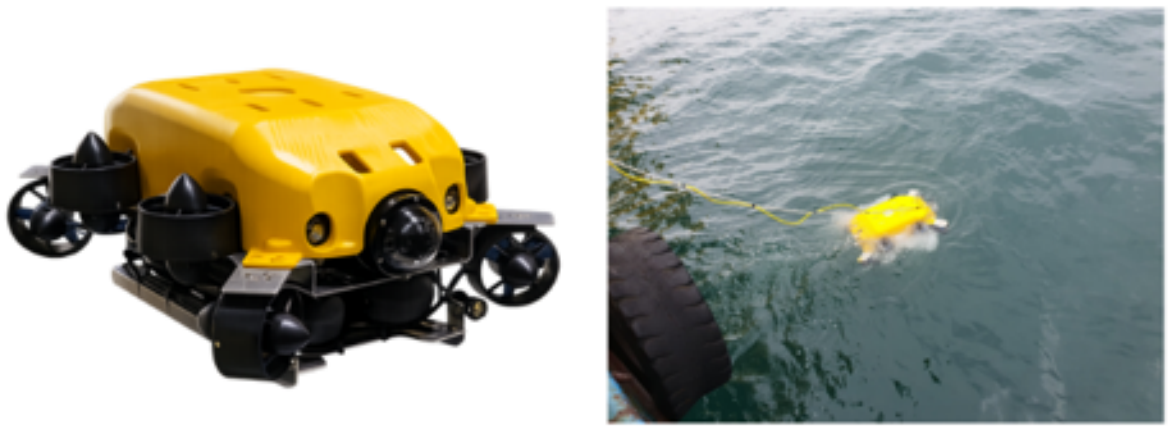

Fig. 6. Underwater robot platform and data acquisition environment

## V. Experiments and Results

This section validates the real-world applicability and performance of the proposed image processing techniques based on image data acquired from an underwater robot operating in real sea environments. Although marine structures such as caissons, tetrapods, and concrete structures exist in real underwater environments, as shown in Fig. 5, the shapes and boundaries of structures are frequently not clearly observable due to marine snow and underwater scattering. Consequently, underwater robot-based structure inspection and monitoring tasks face limitations in terms of image visibility, and image processing technologies are required to improve it. The experiments in this study were conducted on real sea data acquired using the underwater robot platform presented in Fig. 6.

5.1 Experimental Setup

The underwater image data used in the experiments were acquired in real coastal waters of Korea using an underwater robot equipped with a camera sensor. The data include underwater environments in which marine snow and suspended particles are frequently observed, and were composed to reflect real underwater observation conditions in which the shapes of structures are difficult to identify. For these data, the processing results are compared qualitatively and quantitatively to evaluate the image enhancement effect in real underwater robot operating environments.

Forward scattering and post-processing parameters. In the basic synthesis stage (Eq. (1)), we used the standard attenuation coefficients $\beta_c$ for the ten Jerlov water types (Open Ocean I, IA, IB, II, III / Coastal 1, 3, 5, 7, 9), and for each water type the depth $d$ was randomly sampled, within the valid range of each water type, from the interval $0.3 \leq d \leq 15.0$ m, and eight synthetic images were generated per input clean image. In the forward scattering stage, the PSF standard deviation parameters in Eq. (3) were set identically for all channels as $\sigma_{0,c} = 0.5$ pixel and $k_c = 0.15$ pixel/m. $\sigma_{0,c}$ is a minimum value that reflects the sub-pixel-level baseline PSF width remaining at close range ($d \to 0$) due to the camera optics and digital sampling, and it was selected so that structures in regions close to the scattering-free condition are not excessively weakened. $k_c$ was determined by qualitatively comparing the synthesis results against the depth-dependent boundary weakening observed in real Korean coastal images, and over the depth sampling range $d \in [0.3, 15]$ m of this study, $\sigma c(d)$ varies by approximately 0.45–2.75 pixels. This range is suitable for reproducing perceptible visibility degradation at long range while preserving near-range structural information, and it is also consistent with the linear approximation based on the small-angle scattering assumption [20]. Meanwhile, since $\beta c$ in Eq. (1) already accounts for wavelength-dependent attenuation, the PSF in Eq. (3) is used as a channel-common model that approximates only the spatial diffusion component of forward scattering. The weighting function parameter in Eq. (6) was set to $\gamma = 0.2\ \mathrm{m}^{-1}$, a value designed such that $\alpha fs(d)$ reaches approximately 63% at $d = 1/\gamma = 5$ m, so that the forward scattering effect becomes substantially reflected from the middle region of the depth sampling range.

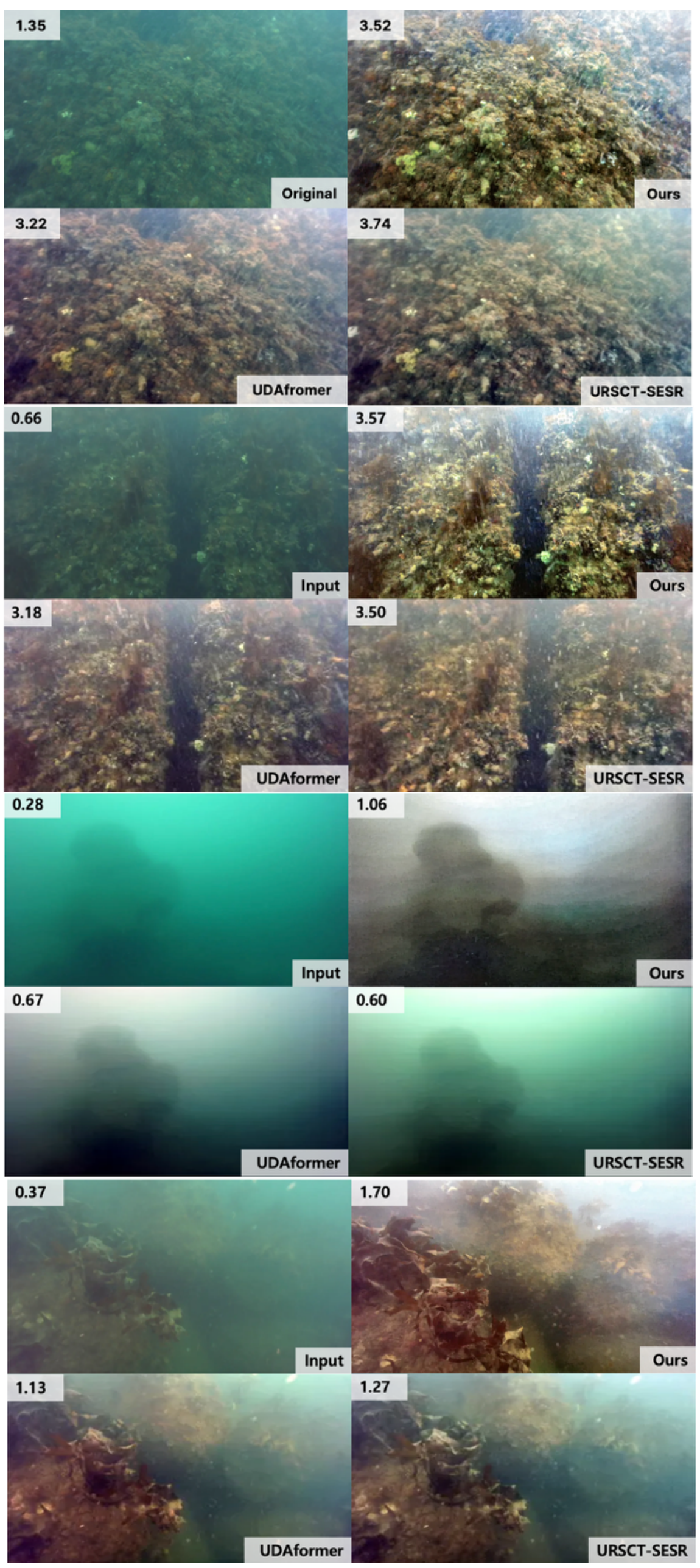

Fig. 7. Qualitative comparison of existing and proposed methods with UIQM scores.

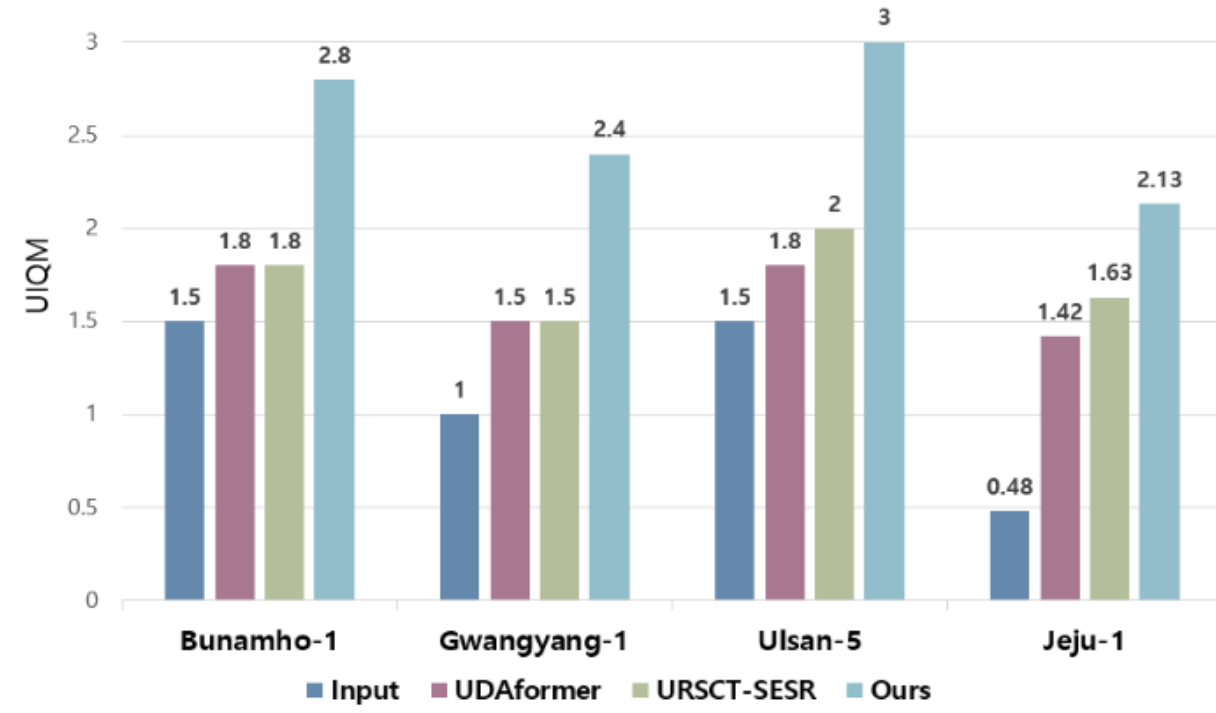

Fig. 8. Quantitative comparison of existing and proposed methods.

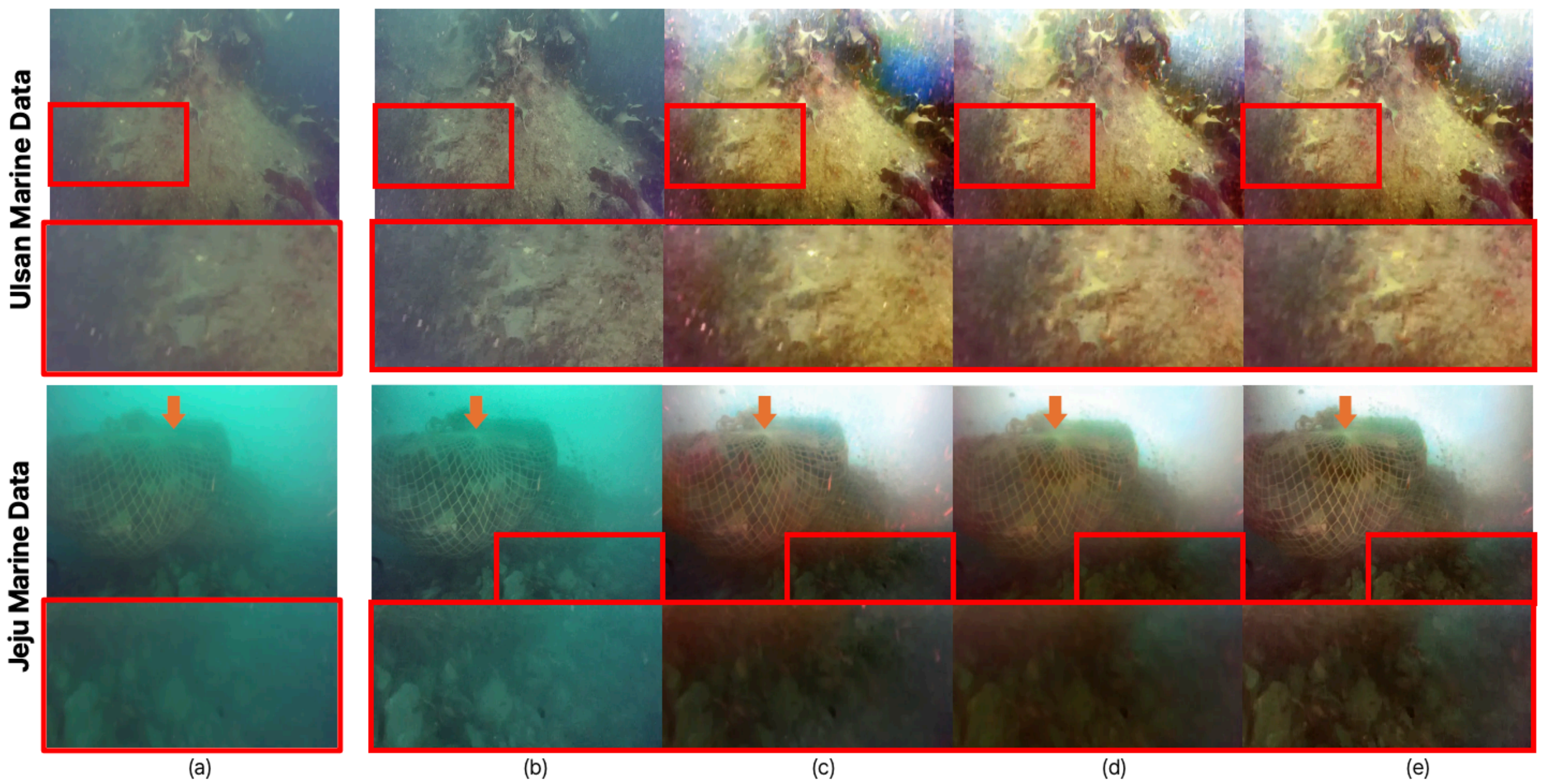

Fig. 9. Qualitative comparison across processing stages of the proposed method. (a) Original RGB input image, (b) result with only histogram equalization and Gaussian Laplacian filtering, (c) result of the original Joint-ID without additional retraining, (d) result of Joint-ID retrained with the proposed synthetic dataset, and (e) final result after applying post-processing to the retrained Joint-ID output.

The marine snow masks were extracted by converting real underwater images to grayscale and binarizing them with an intensity threshold of 70 (on a 0–255 scale), and the Gaussian smoothing standard deviation $\rho$ in Eq. (8) was randomly sampled within the range $0.2 \leq \rho \leq 1.0$. During synthesis, the number of particles per image was set within $1000 \leq N_p \leq 1600$, and the resize scale of each particle was randomly selected within $0.3 \leq s \leq 0.6$. In the post-processing stage (Eq. (9)), the LoG filter standard deviation $\sigma = 1.0$ pixel and the edge enhancement weight $\lambda = 0.3$ were used.

### 5.2 Qualitative and Quantitative Evaluation on Real Sea Data

This subsection presents the qualitative and quantitative evaluation results of the proposed image processing techniques on underwater image data acquired in real sea environments. For quantitative evaluation, UIQM [22], a no-reference underwater image quality metric, was used. UIQM is defined as a linear combination of UICM, UISM, and UIConM, which reflect colorfulness, sharpness, and contrast, respectively, and is expressed as follows.

$$UIQM = c_1 \cdot UICM + c_2 \cdot UISM + c_3 \cdot UIConM \qquad (10)$$

Here, UICM, UISM, and UIConM are the terms reflecting colorfulness, sharpness, and contrast, respectively, and this study uses the weight settings presented in the original paper [22] as they are. Although UIQM is a useful metric for evaluating the overall visual quality of underwater images, it has limitations in directly reflecting individual degradation factors such as marine snow removal performance or the degree of structural boundary restoration. Therefore, in addition to the UIQM-based quantitative evaluation, this study also presents qualitative comparisons to examine how clearly the shapes and boundaries of structures can be identified in real underwater working environments. Fig. 7 shows the qualitative comparison results with existing methods, Fig. 8 shows the quantitative comparison results based on the average UIQM over the entire real sea dataset, and Fig. 9 shows the qualitative comparison of the effects of each application stage of the proposed method.

#### 5.2.1 Comparison with Existing Methods

To compare the performance of the proposed method with existing underwater image processing techniques, the UDAformer [8] and URSCT-SESR [9] models were selected as comparison targets. All comparisons were performed by analyzing the results for the same input images.

UDAformer [8] and URSCT-SESR [9] are methods that mainly focus on dehazing and color restoration of underwater images, and they are effective in improving overall brightness and contrast. In real underwater working environments, however, what matters more than simple color correction or visual sharpness enhancement itself is that the boundaries and shapes of distant structures weakened by forward scattering and marine snow must remain reliably identifiable to operators. In this respect, existing methods have limitations in sufficiently reflecting the complex particle-induced degradation and distance-dependent boundary diffusion that appear in real sea environments.

In the qualitative comparison, the existing methods were effective in enhancing image contrast, but tended to be unstable in restoring the boundaries and detailed structural information of marine structures located in distant regions. In contrast, the proposed method maintained the shapes and boundaries of marine structures relatively stably across the entire image while improving brightness and visibility in a balanced manner, and consequently provided results in which even structures located in distant regions could be identified more clearly.

In the quantitative evaluation as well, the proposed method recorded high UIQM scores under various sea conditions, demonstrating superior performance in improving the visual quality of underwater images.

5.2.2 Stage-wise Qualitative Evaluation of Each Component

Fig. 9 qualitatively compares the results obtained by sequentially applying the processing stages and training conditions of the proposed method to the same input images. In each result image, boxes are marked to highlight regions where marine snow is dense or structural boundaries are included, and these regions are enlarged and presented together in the result images.

Fig. 9(a) is the original RGB image that serves as the reference for comparison.

Fig. 9(b) is the result of applying only histogram equalization and the Laplacian of Gaussian filter, without any learning-based model. Contrast is partially improved, but no clear improvement is observed in terms of structure restoration or visibility enhancement.

Fig. 9(c) is the result of applying the original Joint-ID model without additional training. For the data from the Ulsan waters, fine dot-like particle patterns formed by marine snow are still observed in the boxed regions, and for the data from the Jeju Island waters, although overall color restoration is achieved, the boundaries of the seabed structures and the net structure in the boxed regions remain blurred, so that structural details are not clearly distinguished.

Fig. 9(d) is the result of retraining the Joint-ID model with the synthetic dataset generated by incorporating forward scattering and marine snow artifacts. In the Ulsan waters, the scattered particle artifacts formed by marine snow are clearly reduced compared with (c), particularly around the boxed regions, whereas in the Jeju Island waters, learning-based restoration alone does not sufficiently enhance the boundary contrast of the seabed structures, so the change in structural visibility remains limited.

Fig. 9(e) is the result of additionally applying histogram equalization and the Laplacian of Gaussian filter to the output of the retrained Joint-ID. In the Jeju Island waters, the boundary contrast of the artificial structures distributed on the seabed and of the net structure indicated by the arrow is improved compared with (d), particularly around the boxed regions, so that fine contours are clearly distinguished. This shows that the filter-based post-processing contributes to emphasizing the boundaries of seabed structures, which was limited with learning-based restoration alone.

## VI. Conclusion

This paper pointed out the limitation that existing synthetic underwater images do not sufficiently reflect the degradation caused by forward scattering and marine snow, and proposed a realistic synthetic underwater data generation technique to compensate for this. Forward scattering was approximately modeled using depth-dependent blur, and a synthetic dataset incorporating marine snow patterns extracted from real underwater images was constructed and applied to the Joint-ID model. Through experiments on real underwater robot images from Korean coastal waters, we confirmed that the proposed synthetic data improve visual quality and UIQM performance.

In future work, we plan to apply the proposed method to various underwater image restoration models to verify its generality, and to investigate lightweight designs and processing efficiency improvements for real-time underwater robot operation.

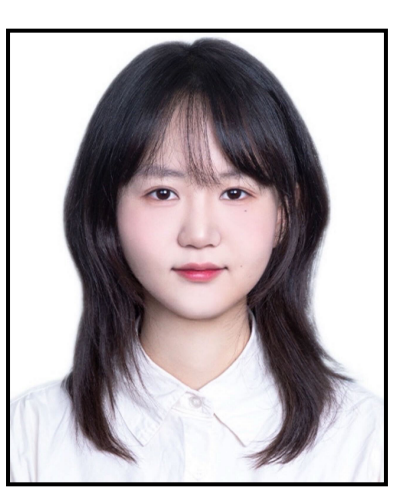

**Seunghee Yun**
2025 B.S. in Mechanical Engineering, Kookmin University. 2025-present: M.S. student, Department of Electrical and Computer Engineering, Inha University. Research interests include sensor fusion and object detection.

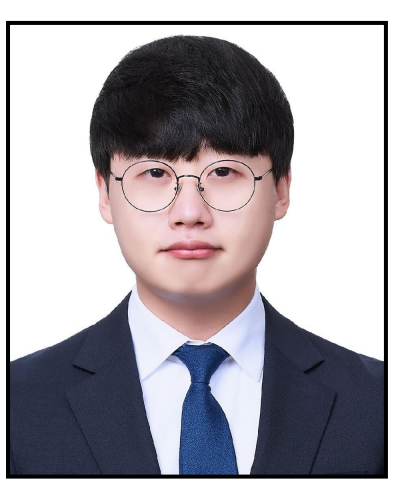

**Geonmo Yang**
2022 B.S. in Electronic Engineering, Tech University of Korea.
2022-present: Integrated M.S.-Ph.D. student, Department of Electrical and Computer Engineering, Inha University. Research interests include SLAM, robust sensing, and long-term autonomy.

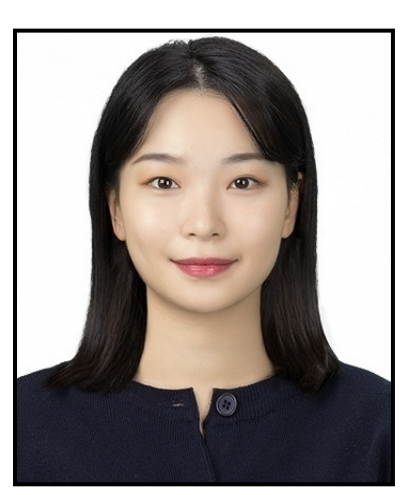

**Juhui Lee**
2022 B.S. in Electrical Engineering, Inha University. 2024 M.S., Department of Electrical and Computer Engineering, Inha University. 2024-present: Ph.D. student, Department of Electrical and Computer Engineering, Inha University. Research interests include 3D vision and visual SLAM.

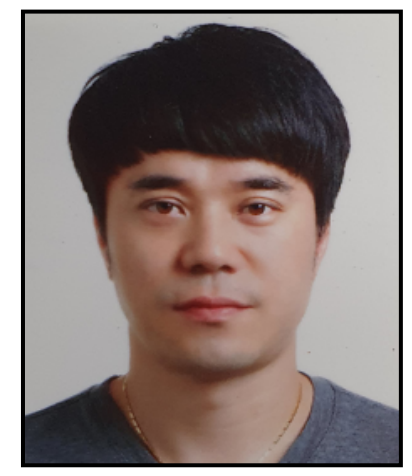

**Changbeom Park**
2008 B.S. in Civil Engineering, Konkuk University. 2010 M.S. in Civil Engineering, Konkuk University. 2010-present: Researcher, R&D Division, Hyundai Engineering & Construction. Currently working on a national R&D project on unmanned underwater construction methods for creating subsea spaces.

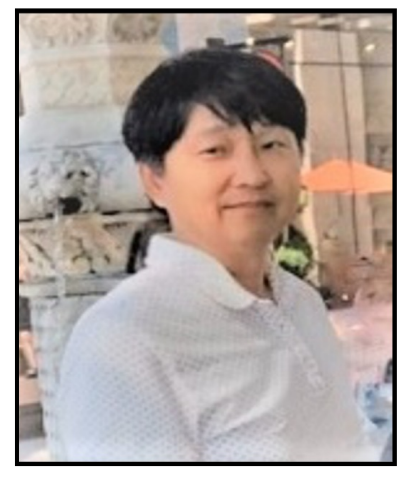

**Jeahyung Choi**
1999 Ph.D., University of Michigan. 2008-present: Researcher, R&D Division, Hyundai Engineering & Construction. Currently working on a national R&D project on unmanned underwater construction methods for creating subsea spaces.

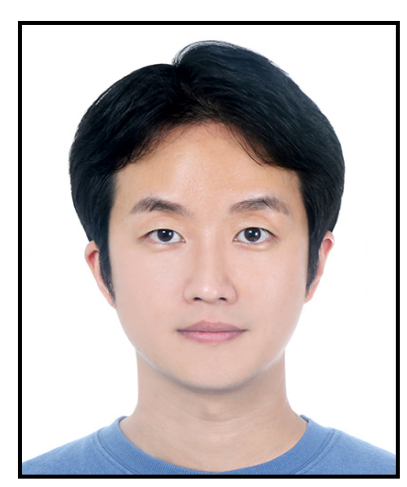

**Younggun Cho**
2013 B.S. in Electrical Engineering, Inha University. 2015 M.S. in Civil and Environmental Engineering, KAIST. 2020 Ph.D. in Civil and Environmental Engineering, KAIST. Currently an assistant professor in the Department of Electrical Engineering, Inha University. Research interests include robust sensing, long-term navigation, and visual localization and mapping.